\documentclass{article}

\PassOptionsToPackage{numbers, compress}{natbib}
\usepackage[final]{nips_2016}

\usepackage[utf8]{inputenc} % allow utf-8 input
\usepackage[T1]{fontenc}    % use 8-bit T1 fonts
\usepackage{hyperref}       % hyperlinks
\usepackage{url}            % simple URL typesetting
\usepackage{booktabs}       % professional-quality tables
\usepackage{amsfonts}       % blackboard math symbols
\usepackage{nicefrac}       % compact symbols for 1/2, etc.
\usepackage{microtype}      % microtypography
\usepackage{graphicx}
\title{Detection of Tooth caries in Bitewing Radiographs using Deep Learning}

\author{
  Muktabh Mayank Srivastava\thanks{The authors contributed equally}\ \\
  ParallelDots, Inc.\thanks{www.paralleldots.xyz}\\
  \texttt{muktabh@paralleldots.com} \\
  \And
  Pratyush Kumar\textsuperscript{*} \\\
  ParallelDots, Inc. \\
  \texttt{pratyush@paralleldots.com} \\
  \AND
  Lalit Pradhan\textsuperscript{*}\\
  ParallelDots, Inc.\\
  \texttt{lalit@paralleldots.com} \\
  \And
  Srikrishna Varadarajan\\
  ParallelDots, Inc.\\
  \texttt{srikrishna@paralleldots.com} \\
}

\begin{document}

% \nipsfinalcopy is no longer used

\maketitle
\begin{abstract}
We develop a Computer Aided Diagnosis (CAD) system, which enhances the performance of dentists in detecting wide range of dental caries. The CAD System achieves this by acting as a second opinion for the dentists with way higher sensitivity on the task of detecting cavities than the dentists themselves. We develop annotated dataset of more than 3000 bitewing radiographs and utilize it for developing a system for automated diagnosis of dental caries. Our system consists of  a deep fully convolutional neural network (FCNN) consisting 100+ layers, which is trained to mark caries on bitewing radiographs. We have compared the performance of our proposed system with three certified dentists for marking dental caries. We exceed the average performance of the dentists in both recall (sensitivity) and F1-Score (agreement with truth) by a very large margin. Working example of our system is shown in Figure~\ref{figured}.
\end{abstract}

\begin{figure}[!ht]
  
  \centering
%  \fbox{\rule[-.5cm]{0cm}{4cm} \rule[-.5cm]{4cm}{0cm}}
  \includegraphics[width=6cm]{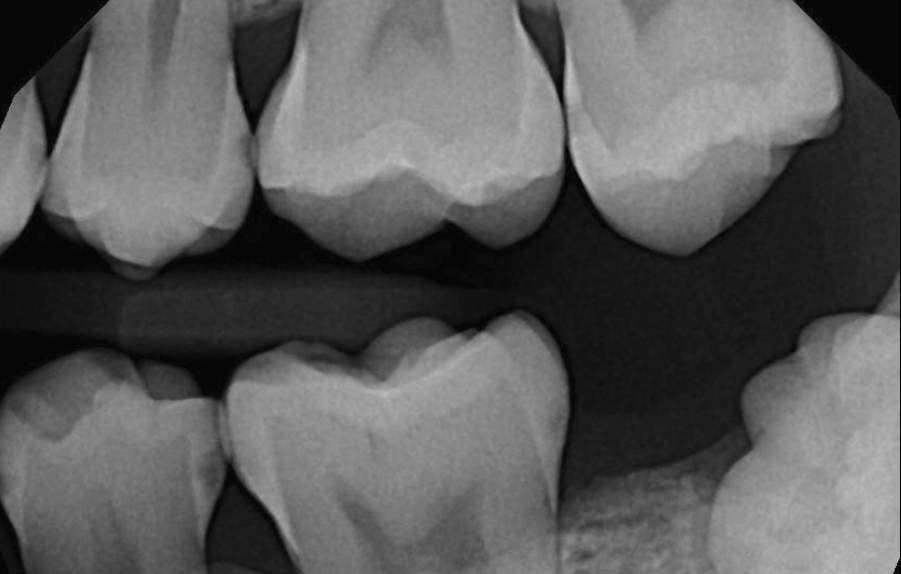} 
  \includegraphics[width=6cm]{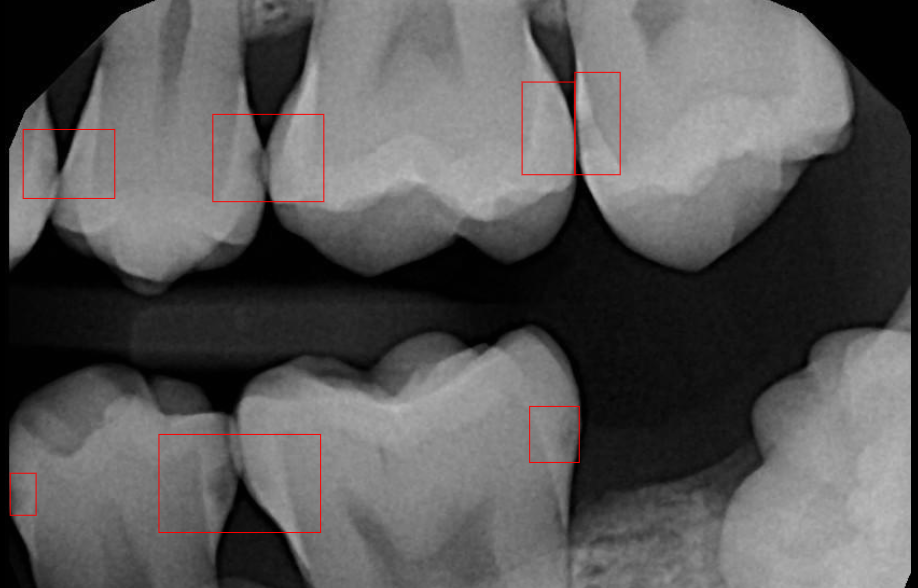}
  \caption{Left: A representative bitewing radiograph. Right: Bounding box in red color   representing caries predicted by our system}
  \label{figured}
\end{figure}

\section{Introduction}

Dental caries (also known as Dental Cavities) are major oral health problem in most industrialized countries, affecting more than one-fourth of U.S. children aged two to five, half of those aged between twelve to fifteen, and more than 90 percent of U.S. adults over age 40 \cite{6stats}. Consequently, dental services (including diagnosis, prevention, and treatment of the diseases related to oral cavity) are among the fastest growing sectors in the healthcare industry \cite{7stats2}.
Dentists often use bitewing radiographs as assisting tools to locate dental caries. This, however, is a challenging task. Dentists rely on clinical experience and patient’s medical history as additional information for corroborating their caries findings on radiographs. Even experienced dentists miss cavities with high probability (20\%-40\%) \cite{2ncbi}, if presented with just the bitewing radiographs. Further, intra- as well as inter-examiner agreement between dentists are very low \cite{3sage}. Automating the dental caries detection process has huge potential in raising standards of medical care by providing increased efficiency and reliability.
Our contributions in the present work are as follows:

\begin{itemize}
\item We develop a large annotated dataset from clinically verified bitewing radiographs
\item We develop a system for automated detection of dental caries from bitewing radiographs
\item We benchmark the efficiency of our proposed system with three practising dentists
\end{itemize}

We develop a System that has better agreement (F1-Score)  with clinically verified data as compared to 3 dentists we work with during our study. On top of this the sensitivity of the system is higher than dentists as well. The study was approved by IRB.

\section{Related Work}

Machine learning is a discipline within computer science that focuses on teaching machines to detect patterns in the underlying data \cite{4bishop}. Machine learning techniques have been previously leveraged for a variety of object detection tasks in natural images. For the task of dental caries detection, very less research work exists that uses traditional machine learning methods. Such traditional machine learning methods are limited by their limited modelling capacities, reliance on feature engineering, and ability to work under specialized non-clinical environments (e.g. detection in extracted tooth artificially arranged for the process). The proposed system, however, is an end-to-end solution, which detects dental cavities directly from the original radiographs without needing any specialized tailoring of images. This has been possible by using methods of Deep learning, specifically a type of Convolutional Neural Networks \cite{5cs231} called Fully Convolutional Neural Networks (FCNN) \cite{1unet}.

\section{Methods}

\subsection{Data Set}
We obtained over 3000 bitewing radiographs from approximately 100 clinics across USA after approval from IRB. All the radiographs were annotated by certified dentists after clinical verification for existence of dental caries. These annotations were used as ground truth for both training as well as testing.
We used 2500 radiographs for training our system. The remaining 500 radiographs were used for testing. These 500 radiographs used for testing were given for marking caries to three practising dentists (henceforth, referred to as testing dentists), who were unaware of the clinical history of the patient. The markings of testing dentists and our system were compared against the ground truth.

\subsection{Development of the System}
Dental cavities appear in amoeboid shapes in the bitewing radiographs, which is quite difficult to annotate. Therefore, we obtained the annotations as loose polygons around the caries. While the loose markings is used to train the System to localize caries, the performance of system is evaluated as a search module for caries.

The caries detection task is trained as a dense classification task which takes as input a bitewing radiograph 2-d image and outputs a binary label (0 or 1) for each pixel. Each output label corresponds to a pixel of the input being caries or not. We train a 100+ layers FCNN to learn the dense classification task for dental cavities in the bitewing radiographs. The threshold for classifying a pixel to be 0 or 1 is obtained qualitatively from the training set results. 5\% of the train set is used as validation set to decide a good fit. No other hyperparameter search was performed.

Since evaluating the System's performance on dense classification metrics like dice coefficient is redundant due to approximate annotations in dataset, we evaluate its performance as a search module for cavities. A tight bounding box is created around the dense classification output of the FCNN and that is treated as the final output of the System. The final output is said to be a successful cavity search if the Jaccard index \cite{8jaccard} (also known as Intersection over Union, IoU) of it is greater than 0.8 with respect to ground truth polygon. A similar criterion is followed to evaluate the polygons marked by testing dentists.

\section{Results}
We report the results as Recall (Sensitivity), Precision (positive predictive value) and Agreement with truth (F1-score). Recall in this case measures ratio of number of cavities successfully searched by the System (or testing dentists) to the total number of cavities in ground truth. Precision is the ratio of number of cavities successfully searched by the System (or testing dentists) to the total number of cavities (succesful and unsuccesful) searched by them. F1-Score of the System (or testing dentists) is the harmonic mean of their Precision and Recall.

The precision, recall and F1-score of our system along with the testing dentists is tabulated in Table~\ref{sample-table}. The high recall of our System shows that it missed only few caries from the ground truth while the dentists are prone to miss a lot more. The relatively low precision of our System shows that there are more false positives but these also consist of ambiguos patches which can be interpreted as potential caries in some cases.  The results clearly demonstrate that our system outperforms the dentists in both sensitivity in predicting caries as well as F1-score by a large margin.

% \begin{center}
% \begin{tabular}{|l|l|l|l|l|}
% Metrics     & System    & Dr. 1		& Dr. 2		& Dr. 3 \\
% Recall 		& 77  		& 45.6     	& 14.2		& 0 \\
% Precision     & 61 		& 57.7      & 100		& 0 \\
% \end{tabular}
% \end{center}

\begin{table}[htbp]
  \caption{Performance comparison between our system and testing dentists}
  \label{sample-table}
  \centering
  \begin{tabular}{|l||*{4}{c|}}\hline
    &\makebox[3em]{System}&\makebox[3em]{Dr. 1}&\makebox[3em]{Dr. 2}
    &\makebox[3em]{Dr. 3}\\\hline\hline
    Recall &80.5 &47.7 &43 &34.4\\\hline
    Precision &61.5 &63 &81.5 &89.1\\\hline
    F1-Score &70 &54 &56 &50 \\\hline
   \end{tabular}
\end{table}

\section{Conclusion}
The use of Computer Aided Diagnosis (CAD) System for clinical diagnosis provides improved performance and reliability along with avoiding  problems caused by intra- and inter- examiner variations. We have presented a System, which automatically finds dental caries in bitewing radiographs. We benchmark the performance of our system against three practicing dentists. Based on the results, we can conclude that our System achieves optimal performance on finding dental caries in bitewing radiographs.

\bibliography{references}

\end{document}